\documentclass[review]{elsarticle}

\usepackage{lineno,hyperref}
\modulolinenumbers[5]

\usepackage{microtype}
\usepackage{graphicx}
\usepackage{booktabs} 
\usepackage[utf8]{inputenc}
\usepackage[english]{babel}
\usepackage{times}
\usepackage{epsfig}
\usepackage{amsmath}
\usepackage{amsthm}
\usepackage{color}
\usepackage{xcolor}
\usepackage{amssymb}
\usepackage{multirow}
\usepackage[normalem]{ulem}
\usepackage[caption=false,font=footnotesize]{subfig}
\usepackage{amsmath,amssymb} 
\usepackage[linesnumbered]{algorithm2e}
\usepackage{float}
\usepackage{tikz}
\usepackage[linesnumbered]{algorithm2e}

\usepackage[normalem]{ulem}

\SetKwInput{KwInput}{Input}                
\SetKwInput{KwOutput}{Output}              
\SetKwRepeat{Do}{do}{while}

\journal{Pattern Recognition}

\begin{document}

\begin{frontmatter}

\title{ANNE: Adaptive Nearest Neighbors and Eigenvector-based Sample Selection for Robust Learning with Noisy Labels}

\author[ufrpe, pa]{Filipe R. Cordeiro\corref{mycorrespondingauthor}}
\ead{filipe.rolim@ufrpe.br}

\author[adelaide]{Gustavo Carneiro}

\address[ufrpe]{Visual Computing Lab, Department of Computing, Universidade Federal Rural de Pernambuco, Brazil}
\address[adelaide]{Centre for Vision, Speech and Signal Processing, University of Surrey, United Kingdom}

\begin{abstract}
An important stage of most state-of-the-art (SOTA) noisy-label learning methods consists of a sample selection procedure that classifies samples from the noisy-label training set into noisy-label or clean-label subsets.
The process of sample selection typically consists of one of the two approaches: loss-based sampling, where high-loss samples are considered to have noisy labels, or feature-based sampling, where samples from the same class tend to cluster together in the feature space and noisy-label samples are identified as anomalies within those clusters.
Empirically, loss-based sampling is robust to a wide range of noise rates, while feature-based sampling tends to work effectively in particular scenarios, e.g., the filtering of noisy instances via their eigenvectors (FINE) sampling exhibits greater robustness in scenarios with low noise rates, and the K nearest neighbor (KNN) sampling mitigates better high noise-rate problems. 
This paper introduces the Adaptive Nearest Neighbors and Eigenvector-based (ANNE) sample selection methodology, a novel approach that integrates loss-based sampling with the feature-based sampling methods FINE and Adaptive KNN to optimize performance across a wide range of noise rate scenarios.
ANNE achieves this integration by first partitioning the training set into high-loss and low-loss sub-groups using loss-based sampling.
Subsequently, within the low-loss subset, sample selection is performed using FINE, while the high-loss subset employs Adaptive KNN for effective sample selection.
We integrate ANNE into the noisy-label learning state of the art (SOTA) method SSR+, and test it on CIFAR-10/-100
(with symmetric, asymmetric and instance-dependent noise), Webvision and ANIMAL-10, where our method shows better accuracy than the SOTA in most experiments, with a competitive training time. The code is available at https://github.com/filipe-research/anne. 
\end{abstract}


\end{frontmatter}


\section{Introduction}\label{sec:introduction}

Deep neural networks have demonstrated remarkable performance in tackling various computer vision applications~\cite{gad2018practical, esteva2021deep, chai2021deep}. To achieve high generalization capacity, these models require a large amount of well-curated training data sets with high-quality labels~\cite{bansal2022systematic}. However, in real-world problems, the annotation of large-scale data is costly and time-consuming. Although crowd-sourcing~\cite{liu2017online} can reduce the laborious labelling process, it may produce low-quality labels due to human annotation mistakes. Moreover, in challenging domains like medicine, labelling can be subject to ambiguity among specialists, making precise and consistent annotations critical for model accuracy.
The main issue of errors in the annotation process is that incorrect labels in the training set can severely harm training, leading to a significant decrease in model generalization~\cite{frenay_classification_2014}. In fact, Zhang et al.~\cite{zhang_understanding_2017} showed that deep learning models can easily memorize label noise, and this process is independent of the model architecture or optimization method~\cite{mentornet}.

Training strategies for learning with noisy labels~\cite{song2022learning} aim to alleviate the adverse effects of label noise using several strategies that usually depend on a sample selection method to subdivide the training set into clean and noisy-label subsets~\cite{dividemix, selfie}. 
After such subdivision, the noisy-label subset is either discarded~\cite{zhang2018generalized}, relabeled~\cite{ssr} or used in a semi-supervised learning approach~\cite{dividemix, longremix}. 
A particularly simple and successful sample selection strategy, which works effectively for a wide range of label noise rates, is the loss-based selection that assumes that small-loss samples have clean labels and high-loss samples have noisy labels~\cite{wei2020combating, dividemix, coteaching}. 
An alternative strategy is based on feature-based sampling, where samples belonging to the same class typically form clusters in the feature space, with samples carrying noisy labels recognized as anomalies within these clusters~\cite{fine,ssr}.
As an example of feature-based sampling, the filtering noisy instances via their eigenvectors (FINE)~\cite{fine} method discriminates clean and noisy-label samples via the distance
to class-specific eigenvectors. In this technique, clean-label samples tend to lie closer to the class-specific dominant eigenvector of the latent representations than the noisy-label samples. FINE is highly effective in low noise-rate scenarios, but tends to degrade in high-noise rate scenarios~\cite{fine}. 
Conversely, SSR+~\cite{ssr} incorporates a sample selection criterion using $K$ nearest neighbor (KNN)~\cite{larose2014k} classification in the feature space. While effective in high noise-rate scenarios, SSR+ exhibits limitations in low-noise situations~\cite{ssr}.

\begin{figure}
    \centering
    \includegraphics[width=1\linewidth]{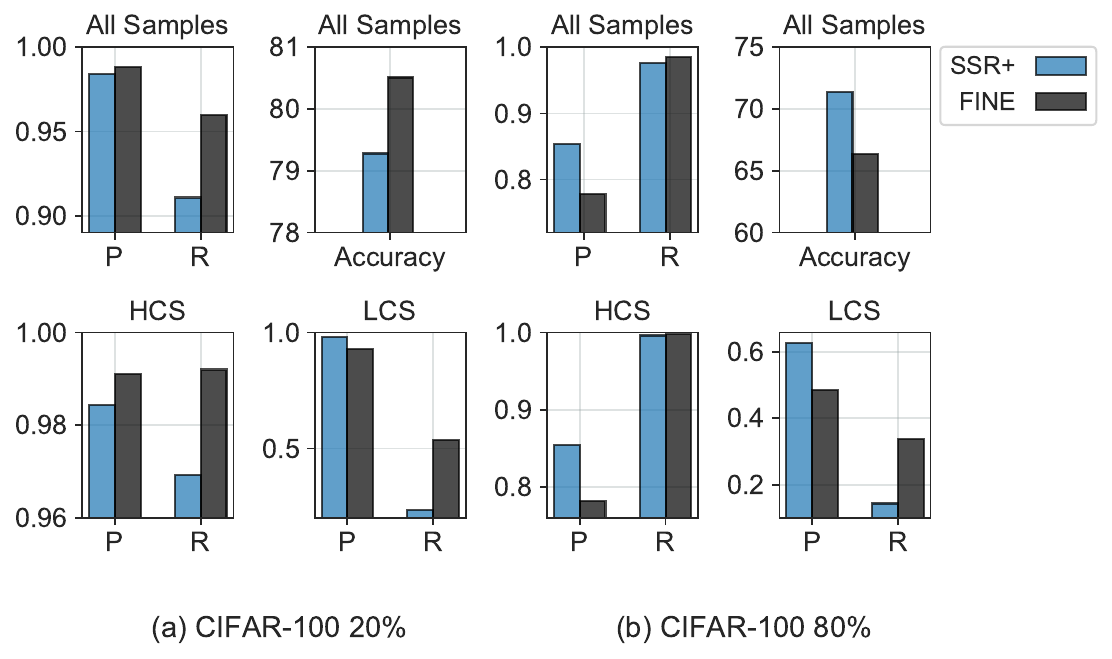}
    \caption{Mean classification accuracy, Precision (P), and Recall (R) of the detection of clean or noisy-label samples done by SSR+ (using KNN)~\cite{ssr} and FINE~\cite{fine} on the last 10 epochs (out of 300) on CIFAR-100 with symmetric noise rate of 20\% (a) and 80\% (b).
    The top row shows the P and R of the clean or noisy-label sample detection, together with the CIFAR-100 classification accuracy for SSR+ and FINE.
    The bottom row shows the P and R of the clean or noisy-label classification for 
    the High Confidence Set (HCS) and Low Confidence Set (LCS), formed with the  small-loss selection from DivideMix~\cite{dividemix}.
    On the top row, note that FINE works better (higher accuracy, P and R) for the low-noise rate scenario, while SSR+ is better for high-noise rate (higher accuracy and P, but comparable R).
    On the bottom row, FINE shows better sample selection results  for the HCS samples of both datasets, while SSR+ is better for the LCS samples from both datasets.    
    }
    \label{fig:motivation}
\end{figure}

Existing noisy-label learning methods in the literature apply just a single sample selection strategy (e.g., based on small loss hypothesis, Eigenvector decomposition, or KNN)~\cite{song2022learning}. 
Our hypothesis is that by dynamically adapting the selection criterion during the training process and leveraging loss-based selection together with FINE and KNN, we can design a more efficacious sample selection strategy.
The top row of Fig.~\ref{fig:motivation} shows an analysis of two feature-based selection strategies, SSR+ (using KNN)~\cite{ssr} and FINE~\cite{fine}, on CIFAR-100 on low (i.e. 20\%) and high (i.e. 80\%) symmetric noise rate.
In that figure, we measure the Precision (P) and Recall (R) of the detection of clean or noisy-label training samples, and the classification accuracy.
The bottom row of Fig.~\ref{fig:motivation}  shows P and R of the detection of clean or noisy-label training samples on the high confidence subset (HCS) and low confidence subset (LCS), built using the small loss strategy~\cite{dividemix}, where HCS contains samples with low loss (i.e., samples classified as clean) and LCS has samples with high loss (i.e., samples classified as noisy).
In the top row of Fig.~\ref{fig:motivation}, we note that FINE has higher accuracy, P and R for the low-noise rate scenario, while SSR+ (using KNN) has higher accuracy and P, but comparable R for the high-noise rate problem.
In the bottom row of Fig.~\ref{fig:motivation}, we note that FINE tends to perform better than SSR+ for the HCS samples in both datasets, while SSR+ appears to be better for the LCS samples in both datasets, so 
we hypothesise that FINE should be used for the HCS samples, while SSR+ should be applied to the LCS samples.

This paper proposes the Adaptive Nearest Neighbors and Eigenvector-based (ANNE) sample selection strategy for robustly learning with noisy labels. 
In this strategy, we first subdivide the training samples into HCS and LCS using a loss-based selection method~\cite{dividemix}.
Subsequently, HCS samples are sub-divided into clean and noisy using FINE, while the remaining LCS samples are sub-divided into clean and noisy using the proposed Adaptive $k$-nearest neighbours (AKNN) method. 
The AKNN method adjusts the value of $K$ based on the neighbour density in the feature space. The combination of  FINE and AKNN is designed to be highly robust to both low and high noise rates. 
To the best of our knowledge, this is the first approach that investigates the  combination of multiple sample selection strategies  to split clean and noisy-label samples. Our selection method is integrated into the iterative selection and relabeling training process of the Sample Selection and Relabelling (SSR+) method~\cite{ssr}, which is a state-of-the-art (SOTA) noisy-label learning method based on sample selection. 
To summarise, the main contributions of the proposed method are as follows:
\begin{enumerate}
    \item A new sample selection methodology for noisy-label learning methods,  leveraging the benefits of employing different sample selection strategies, based on small-loss hypothesis, adaptive K nearest neighbours (AKNN), and Eigenvector decomposition (FINE);
    \item An adaptive $k$-nearest neighbor approach for selecting low-confidence clean-label samples, by dynamically adjusting the $k$-value based on neighborhood density, to make the approach robust to both high and low label noise rate scenarios;
    \item Incorporation of ANNE into SSR+~\cite{ssr}, which is a SOTA noisy-label learning method based on sample selection, allowing us to validate our methodology  on real-world datasets. 
\end{enumerate}
Empirical results on CIFAR-10/-100~\cite{cifar} under symmetric, asymmetric and instance-dependent
noise, show that ANNE outperforms previous SOTA approaches. For real-world data sets, like Webvision, ANNE presents the best results in the field. Our results also show that ANNE has a competitive training time compared to SOTA approaches. 

\section{Prior Work}

Current noisy-label learning methods available in the literature~\cite{song2022learning} usually explore two types of sample selection strategies, namely: loss-based sampling~\cite{mentornet} and feature-based sampling~\cite{ssr}. 
Loss-based methods operate on the premise that clean samples are generally associated with lower loss values particularly during the first training stages. 
Such sample selection methods strongly rely on a loss threshold that can be defined in many ways, such as with  a fixed value~\cite{mentornet, shen2019learning, longremix}, or with an adaptive value using a Gaussian Mixture Model (GMM)~\cite{dividemix} or Otsu~\cite{kaiser2022blind}. 
Also exploring loss-based sampling techniques, Co-teaching~\cite{coteaching} uses a peer network to select the samples by agreement, using small-loss instances. DivideMix~\cite{dividemix} combines Co-teaching with the low-loss strategy for sample selection, using GMM to find the optimal loss threshold. After classifying the samples into clean and noisy-label subsets, the methods will train the model using a semi-supervised learning. 
PropMix~\cite{propmix} introduces a two-stage sample selection process that employs second-order filtering to identify easy and hard noisy samples using GMM. 
Huang et al.~\cite{o2unet} employ a cyclic status transition strategy, tuning their model's hyperparameters to shift between overfitting and underfitting states, where clean samples are selected based on the historical training loss recorded during these cycles.

 In feature-based sampling methods~\cite{ssr, fine, wu2020topological}, both clean and noisy samples are distinguished by analyzing their features. FINE~\cite{fine} proposes a sample selection based on eigenvector decomposition, where clean samples tend to be closer to the class-dominant eigenvector of the latent representations than the noisy-label samples. FINE also needs to define a threshold based on the similarity to the class-dominant eigenvector to identify a sample as clean. Although this strategy is effective for low noise rates, it struggles in high noise scenarios~\cite{fine}.
 Noisy Graph Cleaning (NGC)~\cite{wu2021ngc} iteratively constructs a nearest neighbour graph using latent representations and performs geometry-based sample selection by aggregating information from neighbourhoods, where soft pseudo-labels are assigned to samples classified as having noisy labels. SSR+~\cite{ssr} introduces a selection strategy based on KNN using the feature space, where samples with labels that match the most similar class in a K-sized neighbourhood are classified as clean. In SSR+, the value for K is a fixed hyperparameter for all datasets, but the optimal value may vary depending on the noise rate, as will be shown in Sec.~\ref{sec:results}. 
 SSR+ also employs a relabeling strategy based on the confidence model and, in contrast to recent models, it utilizes a single network. Zhao et al. proposes Centrality and Consistency (CC) method~\cite{centralityc}, which  uses a two-stage sampling procedure, including class-level feature clustering followed by a consistency score. In CC's first stage, they employ a class-level feature-based clustering procedure to identify clean samples according to their cosine similarity to the corresponding class-wise prediction centres. The second stage uses two classifiers to evaluate the prediction consistency.

The main difference between existing sample selection methods and
our approach is that we provide a hybrid adaptive strategy based on 
loss-based sampling and feature-based sampling.  We hypothesise that samples with low and high confidence scores require different sample selection strategies with adaptive parameters. Our approach demonstrates the effectiveness of applying
loss-based sampling to divide the training set into HCS and LCS, where FINE further subdivides the HCS samples, and AKNN subdivides the LCS using adaptive values of K.

\section{Method}
\label{sec:method}

\subsection{Problem Setup }
\label{sec:problem_definition}

We denote the training set by $\mathcal{D}=\{(\mathbf{x}_i, \mathbf{y}_i)\}_{i=1}^{|\mathcal{D}|}$, with  $\mathbf{x}_i \in \mathcal{S} \subset \mathbb{R}^{H \times W \times 3}$ being the $i^{th}$ RGB image of size $H \times W$, and $\mathbf{y}_i \in \{0,1\}^{|\mathcal{Y}|}$ denoting a one-hot vector representing the given label, with $\mathcal{Y} = \{1,...,|\mathcal{Y}|\}$ denoting the set of labels, and $\sum_{c \in \mathcal{Y}} \mathbf{y}_i(c)=1$. 
The hidden true label $\hat{\mathbf{y}}_i$ can differ from the given noisy label  $\mathbf{y}_i$ as a result of the label transition probability represented by 
$p(\mathbf{y}(j) = 1 | \mathbf{x}_i,\hat{\mathbf{y}}_i(c) = 1)=\eta_{jc}(\mathbf{x}_i)$,
where the $j,c\in\mathcal{Y}$ are the classes, $\eta_{jc}(\mathbf{x}_i) \in [0,1]$ the probability of flipping the class $c$ to $j$, and $\sum_{j \in \mathcal{Y}}\eta_{jc}(\mathbf{x}_i)=1$. There are three common types of noise in the literature:  
symmetric~\citep{kim2019nlnl}, asymmetric~\citep{patrini2017making}, and instance-dependent~\citep{rog}.
The symmetric noise is a label noise type where the hidden true labels are flipped to a random class
with a fixed probability $\eta$, where the true label is included into the label flipping options, which means that $\eta_{jc}(\mathbf{x}_i)=\frac{\eta}{|\mathcal{Y}|-1}, \forall j \in \mathcal{Y}, \text{ such that } j \neq c$, and $\eta_{cc}(\mathbf{x}_i)=1-\eta$. 
The asymmetric noise has its labels flipped 
between similar-looking object categories~\citep{patrini2017making}, where $\eta_{jc}(\mathbf{x}_i)$ depends only on the classes $j,c\in\mathcal{Y}$, but not on $\mathbf{x}_i$. 
Finally, the instance-dependet noise~\citep{rog} is the noisy type where the label flipping depends both on the classes $j,c\in\mathcal{Y}$ and image $\mathbf{x}_i$. 

\subsection{ANNE}
\label{sec:method}

\begin{figure}[!ht]
\centering
\includegraphics[ width=\columnwidth]{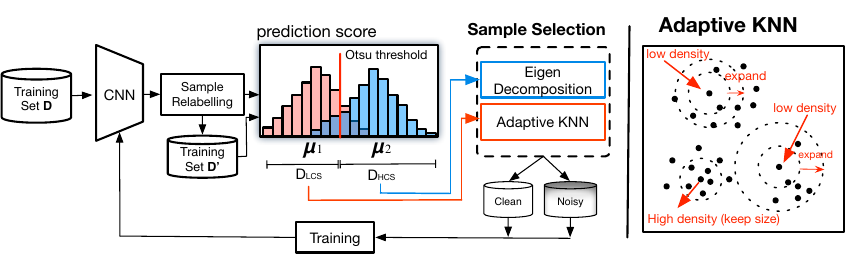}
\caption{Our ANNE sample selection strategy starts with a sample relabeling approach~\cite{ssr}, followed by a sample selection stage that first divides the training set into high-confidence ($\mathcal{D}_{HCS}$) and low-confidence ($\mathcal{D}_{LCS}$)  subsets, according to the classification probability of the samples. Then, the samples from $\mathcal{D}_{LCS}$ will be divided into clean or noisy-label using an adaptive KNN approach, while samples from $\mathcal{D}_{HCS}$ will be divided into clean or noisy-label using an eigen-decomposition technique. As shown in the rightmost figure, our proposed adaptive KNN automatically changes the range to find the $K$ nearest neighbors, depending on the density of the sample in the feature space.}
\label{fig:anne}
\end{figure}

The proposed ANNE method is a sample selection approach that uses loss-based selection together with Eigenvector Decomposition and Adaptive $k$-nearest neighbors strategies.
Fig.~\ref{fig:anne} shows ANNE's framework.

ANNE is composed of a feature encoder $f_\phi: \mathcal{S} \to \mathcal{Z} \subset \mathbb{R}^d$ and a classifier $g_\theta: \mathcal{Z} \to \Delta^{|\mathcal{Y}|-1}$, with $\Delta^{|\mathcal{Y}|-1}$ denoting the probability simplex for the $|\mathcal{Y}|$-class problem. 
ANNE starts with a sample relabeling procedure~\cite{ssr}  
 using a threshold value $\gamma_r$ to threshold the maximum prediction score for each training sample , as follows:
\begin{equation}
\mathbf{y}_i = \left\{\begin{matrix}
 \mathsf{OneHot}(g_{\theta}(f_{\phi}(\mathbf{x}))), & \text{ if } \max (g_{\theta}(f_{\phi}(\mathbf{x}))) > \gamma_r \\ 
\mathbf{y}_i, & \text{ if } \max (g_{\theta}(f_{\phi}(\mathbf{x}))) \leq \gamma_r \\
\end{matrix}\right.,
\end{equation}
where the function $\mathsf{OneHot}(g_{\theta}(f_{\phi}(\mathbf{x})))$ transforms the categorical distribution produced by $g_{\theta}(.)$ into a one-hot vector of size $|\mathcal{Y}|$, with the class containing the maximum probability assigned to $1$, and all other classes assigned to $0$.

Next, ANNE runs the sample selection stage.
Initially, we use the maximum prediction scores to divide the samples into two subsets, leveraging the Otsu's algorithm~\cite{otsu}, as in~\cite{kaiser2022blind}.
The first step is to determine a threshold value, denoted as
$\tau$,  which splits 
$\mathcal{D}$ into two subsets, the high-confidence subset (HCS)  $\mathcal{D}_{HCS} = \{ (\mathbf{x},\mathbf{y}) | (\mathbf{x},\mathbf{y}) \in \mathcal{D} \text{, and } \max(g_{\theta}(f_{\phi}(\mathbf{x}))) \ge \tau \}$ and the low-confidence subset (LCS) $\mathcal{D}_{LCS} = \{ (\mathbf{x},\mathbf{y}) | (\mathbf{x},\mathbf{y}) \in \mathcal{D} \text{, and } \max(g_{\theta}(f_{\phi}(\mathbf{x}))) < \tau \}$.
To find  the optimal threshold $\tau$, we maximize the following objective function:
\begin{equation}
\tau^* = \arg\max_{\tau} \frac{n_{NCS}(\tau)(\mu_{HCS}(\tau)-\mu)^2+ n_{LCS}(\tau)(\mu_{LCS}(\tau)-\mu)^2}{n_{NCS}(\tau)\sigma_{HCS}(\tau)^2+n_{LCS}(\tau)\sigma_{LCS}(\tau)^2},
\label{eq:otsu_optimisation}
\end{equation}
where $n_{NCS}(\tau) = |\mathcal{D}_{HCS}|$ and $n_{LCS}(\tau) = |\mathcal{D}_{LCS}|$ 
, $\mu_{HCS}(\tau)$, $\sigma_{HCS}(\tau)$, $\mu_{LCS}(\tau)$, $\sigma_{LCS}(\tau)$ denote the mean and standard deviation of the maximum classification probabilities in $\mathcal{D}_{HCS}$ and $\mathcal{D}_{LCS}$, respectively, and $\mu$ denotes the mean of the maximum classification probabilities of all samples in $\mathcal{D}$. The optimal value of $\tau$ maximises the inter-class separation, and  minimizes the   variance. 
In practice, we estimate $\tau^*$ in Eq.~\eqref{eq:otsu_optimisation} by varying $\tau$ with a step size of $0.001$, within a range between 0 and 1.

To enable a more effetive selection of clean and noisy samples from the subsets $\mathcal{D}_{HCS}$ and $\mathcal{D}_{LCS}$, we apply a method robust to large noise rate problems to the LCS subset $\mathcal{D}_{LCS}$, and a method robust to small noise rate problems to the HCS subset $\mathcal{D}_{HCS}$.
In particular, 
we use Eigen Decomposition for $\mathcal{D}_{HCS}$, whereas for $\mathcal{D}_{LCS}$ we apply adaptive KNN. 
We empirically evaluate various combinations of strategies for these groups in Sec.~\ref{sec:results}, but the setup described above yields the most robust and consistent results across different noise rates.

\subsubsection{Adaptive KNN (AKNN)}
\label{sec:AKNN}

\begin{figure}
    \centering
    \includegraphics[width=\linewidth]{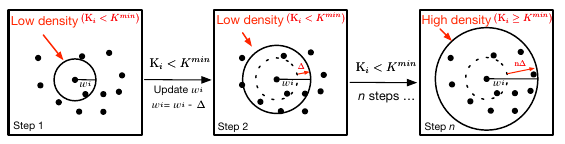}
    \caption{Our AKNN  strategy automatically defines the K value of the K-nearest neighbour based on local density. For each sample $i$, we estimate the number of nearest neighbours $K_i$  based on the number of neighbours with cosine similarity above threshold $\omega_i$. If a low density region is identified (i.e. $K_i < K^{min}$) the threshold $\omega_i$ is reduced by $\Delta$. This process continues until the condition $K_i \ge K^{min}$ is achieved. For samples that initially are already in high-density regions, this process is skipped. }
    \label{fig:aknn}
\end{figure}

We propose an adaptive $K$-nearest neighbour for noisy labels, where the value of $K$ varies according to the local density in the feature space. 
The cosine  similarity between two samples $\mathbf{x}_i$ and $\mathbf{x}_j$  is denoted by
\begin{equation}
s_{ij}=\frac{f_{\phi}(\mathbf{x}_i)^{\top}f_{\phi}(\mathbf{x}_j)}{\left \| f_{\phi}(\mathbf{x}_i) \right \|_2 \times \left \| f_{\phi}(\mathbf{x}_j) \right \|_2},
\label{eq:cosine_similarity}
\end{equation}
where $f_{\phi}(\mathbf{x}_i)$ and $f_{\phi}(\mathbf{x}_j)$ correspond to the features of samples $\mathbf{x}_i$ and $\mathbf{x}_j$. 
Unlike traditional KNN approaches, we propose an adaptive KNN method, where the number of nearest neighbors, denoted by $K_i$, varies depending on the local density of the training sample in the feature space $\mathcal{Z}$. Fig.~\ref{fig:aknn} shows our AKNN framework to automatically define $K_i$.
We initially retrieve the nearest neighbours based on the samples with cosine similarity above threshold $\omega_i$, for each sample $(\mathbf{x}_i,\mathbf{y}_i) \in \mathcal{D}_{LCS}$, forming the set $\mathcal{N}_{\mathbf{x}_i} = \{ (\mathbf{x}_j,\mathbf{y}_j) | (\mathbf{x}_j,\mathbf{y}_j) \in \mathcal{D}_{LCS}, i \neq j, s_{ij} > \omega_i \}$.  
We initially set  $\omega_i$ to a high value and iteratively reduce it, while observing the number of neighbours $K_i=|\mathcal{N}_{\mathbf{x}_{i}}|$ for each sample $(\mathbf{x}_i,\mathbf{y}_i) \in \mathcal{D}_{LCS}$. 
The minimum value of $K_i$ is defined as $K^{min}$, and we control the number of neighbours by decrementing $\omega_i$.  
For each sample $(\mathbf{x}_i,\mathbf{y}_i) \in \mathcal{D}_{LCS}$,  $\omega_i$  is initialized at $0.99$ and, at each iteration where $K_i < K^{min}$ we decrease $\omega_i$ by $\Delta_{s}=0.01$.  
We define $K^{min}$ by further subdividing $\mathcal{D}_{LCS}$ with 
\begin{equation}
\begin{split}
\mathcal{D}_{LCS1} = \{ & 
(\mathbf{x},\mathbf{y}) | (\mathbf{x},\mathbf{y}) \in \mathcal{D} \text{, and } 
\mu_{LCS}(\tau) \leq \max(g_{\theta}(f_{\phi}(\mathbf{x})))  < \tau \}, \\ 
\mathcal{D}_{LCS2} = \{ &  (\mathbf{x},\mathbf{y}) | (\mathbf{x},\mathbf{y}) \in \mathcal{D} \text{, and } \max(g_{\theta}(f_{\phi}(\mathbf{x}))) < \mu_{LCS}(\tau) \},
\end{split}
\label{eq:subsets}
\end{equation}
where $\mu_{LCS}(\tau)$ is the mean of the maximum classification probabilities in $\mathcal{D}_{LCS}$ defined in Eq.~\eqref{eq:otsu_optimisation}, and 
$\mathcal{D}_{LCS1}$ and $\mathcal{D}_{LCS2}$ represent the sets of low and medium-low confidence samples, respectively.
We use different values of $K^{min}$ for  subsets $\mathcal{D}_{LCS1}$ and $\mathcal{D}_{LCS2}$, defined as $K_{LCS1}^{min}$ and $K_{LCS2}^{min}$. 
As the subset $\mathcal{D}_{LCS1}$ and $\mathcal{D}_{LCS2}$ refer to the groups with high noisy-label samples, we use a large value for $K^{min}$, with $K_{LCS1}^{min}=40$ and $K_{LCS2}^{min}=80$. 
Since $\mathcal{D}_{LCS2}$ is expected to be more corrupted than $\mathcal{D}_{LCS1}$ due to the lower prediction scores, we use a higher value for $K_{LCS2}^{min}$.
These values were estimated empirically and remained the same for all experiments. 
Samples with labels matching the prediction from the KNN classifier, denoted by the function $\mathsf{KNN}(\mathbf{x}_i,K_i)$ that returns the most common class in $\mathcal{N}_{\mathbf{x}_{i}}$, are classified as clean, while those samples that do not match the KNN prediction are classified as noisy. 
At the end of the sample selection process, we obtain the following subsets from $\mathcal{D}_{LCS}$: $\mathcal{D}'_{clean}$ and $\mathcal{D}'_{noisy}$, as explained in Algorithm~\ref{alg:ANN} that shows the pseudocode of our adaptive KNN approach.

\RestyleAlgo{ruled}
\begin{algorithm}[t!]
\scriptsize

\KwInput{$\mathcal{D}_{LCS1}$, $\mathcal{D}_{LCS2}$,  $f_\phi(.)$, $K_{LCS1}^{min}$,$K_{LCS2}^{min}$, $\Delta_s$}
$\mathcal{D}'_{clean} = \emptyset$, $\mathcal{D}'_{noisy} = \emptyset$ \\
 \For {$ (\mathbf{x}_i,\mathbf{y}_i) \in \mathcal{D}_{LCS1} \cup \mathcal{D}_{LCS2}$}{
  $\omega_i = 0.99$ \\
  \eIf{ $(\mathbf{x}_i,\mathbf{y}_i) \in \mathcal{D}_{LCS1} $}{
  $K^{min}= K_{LCS1}^{min}$ \\
  }
  {
  $K^{min}= K_{LCS2}^{min}$ \\
  }
  
  \Do {$K_i < K^{min}$}{
    $K_i = |\mathcal{N}_{\mathbf{x}_i}|$, where $\mathcal{N}_{\mathbf{x}_i} = \{ (\mathbf{x}_j,\mathbf{y}_j) |  (\mathbf{x}_j,\mathbf{y}_j) \in \mathcal{D}_{LCS1} \bigcup \mathcal{D}_{LCS2}, i \neq j, s_{ij} > \omega_i \}$\\
     
    $\omega_i = \omega_i - \Delta_s$
  }
  $\mathcal{D}'_{clean} = \mathcal{D}'_{clean} \bigcup (\mathbf{x}_{i}, \mathbf{y}_i)$ if $\mathbf{y}_i = \mathsf{KNN}(\mathbf{x}_i,K_i)$ \\
$\mathcal{D}'_{noisy} = \mathcal{D}'_{noisy} \bigcup (\mathbf{x}_{i}, \mathbf{y}_i)$ if $\mathbf{y}_i \neq \mathsf{KNN}(\mathbf{x}_i,K_i)$ \\
 }
 \KwOutput{$\mathcal{D}'_{clean}, \mathcal{D}'_{noisy}$}
 \caption{Adaptive KNN}
 \label{alg:ANN}
 
\end{algorithm}

\subsubsection{Eigen Decomposition}
\label{sec:FINE}

To identify the clean and noisy-label samples from $\mathcal{D}_{HCS}$, we use the FINE~\cite{fine} algorithm. 
FINE finds clean and noisy-label instances using the square of inner products between the  image features produced by $f_{\phi}(.)$ and the dominant eigenvector computed from the features belonging to the same class. In this manner, we treat the sample as clean if it is
aligned with the most dominant eigenvector, while most of the noisy-label samples tend not to be as well aligned. FINE uses a threshold parameter $\gamma_e$ to select the samples based on such inner product with the eigenvector.  We note that FINE provides more accurate sample selection for low-noise subsets. 
Consequently,  we apply FINE on $\mathcal{D}_{HCS}$, which predominantly contains clean-label samples. 
At the end of this process, we produce $\mathcal{D}''_{clean}$ and $\mathcal{D}''_{noisy}$. 

\subsubsection{Training}

As depicted in Fig.~\ref{fig:anne}, we propose an effective sample selection strategy by combing AKNN (Sec.~\ref{sec:AKNN}) and FINE (Sec.~\ref{sec:FINE}).
 The  selected clean and noisy-label sample sets are defined as $\mathcal{D}_{clean} = \mathcal{D}'_{clean} \cup \mathcal{D}''_{clean} $ and $\mathcal{D}_{noisy} = \mathcal{D}'_{noisy} \cup \mathcal{D}''_{noisy}$. 
 To deal with the possible class imbalance in the selected subset, we over-sample minority classes~\cite{ssr}. 
 In addition, we  perform a linear combination between clean and noisy-label samples, separately, using MixUp~\citep{mixup} data augmentation. 
 In particular, for each subset, we obtain $\hat{\mathcal{D}}_{clean}=\mathsf{MixUp}(\mathcal{D}_{clean},\mathsf{shuffled}(\mathcal{D}_{clean}))$ and $\hat{\mathcal{D}}_{noisy}=\mathsf{MixUp}(\mathcal{D}_{noisy},\mathsf{shuffled}(\mathcal{D}_{noisy}))$. Following~\cite{ssr, propmix}, we oversample $\hat{\mathcal{D}}_{clean}$ to be proportional to $|D_{clean}|+|D_{noisy}|$.

The optimisation to  train the model is defined as follows~\cite{ssr}:
\begin{equation}
\begin{split}
    \theta^{*},\phi^{*} = \arg\min_{\theta,\phi} &-\frac{1}{|\hat{\mathcal{D}}_{clean}|}\sum_{(\mathbf{x},\mathbf{y})\in\hat{\mathcal{D}}_{clean}} \mathbf{y}^{\top}\log g_{\theta}(f_{\phi}(\mathbf{x}))  \\
    &+\frac{1}{|\hat{\mathcal{D}}_{noisy}|}\sum_{(\mathbf{x},\mathbf{y})\in\hat{\mathcal{D}}_{noisy}} \left ( -\frac{h_1(\mathbf{x},\phi)^{\top}h_2(\mathbf{x},\phi)}{\| h_1(\mathbf{x},\phi) \|_2  \| h_2(\mathbf{x},\phi) \|_2} \right ) ,   
\end{split}    
\label{eq:optimisation}
\end{equation}
where 
$h_1(\mathbf{x},\phi) = h_{pred}(h_{proj}(f_{\phi}(a_1(\mathbf{x}))))$ and 
$h_2(\mathbf{x},\phi) = h_{proj}(f_{\phi}(a_2(\mathbf{x})))$, with $h_{pred}(.)$ representing a predictor head, $h_{proj}(.)$ denoting a projector head, and $a_{\{1,2\}}(.)$ being different data augmentation functions. 
The first term in the optimisation in Eq.~\eqref{eq:optimisation} minimises a cross entropy error using the clean-label subset $\hat{\mathcal{D}}_{clean}$, while the second term minimises the cosine distance between the vectors produced from data augmentation of the same sample from the noisy-label dataset $\hat{\mathcal{D}}_{noisy}$.


\section{Experiments}\label{sec:experiments}

\subsection{Data Sets}
\label{sec:datasets}

We conduct our experiments on the data sets  CIFAR-10, CIFAR-100~\citep{cifar}, ANIMAL-10N~\cite{selfie}  and  WebVision~\citep{webvision}. CIFAR-10 and CIFAR-100 have 50k training and 10k testing images of size $32 \times 32$ pixels, where CIFAR-10 has 10 classes and CIFAR-100 has 100 classes and all training and testing sets have equal number of images per classes. 
As CIFAR-10 and CIFAR-100 data sets originally do not contain label noise. By following the literature~\citep{dividemix}, we add the following synthetic noise types: symmetric (with noise rate $\eta \in \{0.2, 0.5, 0.8, 0.9\}$, asymmetric (using the mapping  in~\citep{dividemix, patrini2017making}, with $\eta_{jc}=0.4$). We also evaluate CIFAR-10 and CIFAR-100 with instance dependent noise (IDN), following~\cite{xia2020part}, with noise rates in $\{0.2, 0.3, 0.4, 0.5 \}$. We also evaluate combined open-set and closed-set noises, as used in \cite{evidentialmix,ssr}. The combined benchmark
 is defined by the rate of label noise in the experiment, denoted by $\rho \in \{0.3,0.6\}$, and the proportion of closed-set noise in the label noise, denoted by $\omega \in \{0.5,1\}$.  
The closed-set label noise is simulated by randomly selecting $\rho\times\omega\times100\%$ of the training samples from CIFAR-10, and symmetrically shuffling their label, similarly to the synthetic label noise used in~\cite{dividemix}.
The open-set label noise is simulated by randomly selecting $\rho\times(1-\omega)\times100\%$ of the training images from CIFAR-10 and replacing them with images randomly selected from  CIFAR-100, where a CIFAR-10 label is randomly assigned to each one of these images.

 WebVision is a real-world noisy-label dataset with 2.4 million images collected from the internet, with the same 1000 classes of ILSVRC12~\citep{deng2009imagenet} and images resized  to $256 \times 256$ pixels. It provides a clean test set of 50k images, with 50 images per class. We compare our model using the first 50 classes of the Google image subset, as used in \cite{dividemix, chen2019understanding}.

 ANIMAL-10N~\cite{selfie} is also a real-world noisy-label dataset consisting
of human-annotated images of 10 animal categories,
with 5 pairs of animals having similar appearances, such as
chimpanzee and orangutan. The dataset has an estimated
label noise rate of 8\%, and it comprises 50,000 training
images and 10,000 test images.

\subsection{Implementation}

For CIFAR-10 and CIFAR-100, we used a 18-layer 
PreaAct-ResNet-18 (PRN18)~\citep{he2016identity}  
as our backbone model, following previous works~\citep{dividemix,ssr}. The models are trained with stochastic gradient descent (SGD) with momentum of 0.9, weight decay of 0.0005 and batch size of 128. The learning rate is 0.02 and is controlled by a cosine annealing scheduler. The models are trained for 300 epochs.   We set $\gamma_r$ (relabeling threshold) and $\gamma_e$ (FINE threshold) using the  values listed in Table ~\ref{tab:parameters_cifar}.

\begin{table}[ht]
\centering
\footnotesize
\scalebox{1.0}{
\begin{tabular}{c|cccc|c||cccl}
\toprule
\multicolumn{1}{c}{\textbf{Data set}} & \multicolumn{5}{c}{\textbf{CIFAR-10}} & \multicolumn{4}{c}{\textbf{CIFAR-100}}\\    
\cmidrule(l){1-1}
\cmidrule(l){2-6}
\cmidrule(l){7-10}
\multicolumn{1}{c}{\textbf{Noise type}} & \multicolumn{4}{c}{\textbf{sym.}} & \multicolumn{1}{c}{\textbf{asym.}} &  \multicolumn{4}{c}{\textbf{sym.}} \\
\midrule

parameter/ noise ratio &   20\% & 50\% & 80\% & 90\% & 40\% & 20\% & 50\% & 80\% & 90\%  \\
$\gamma_r$&  0.9 & 0.9 & 0.8 & 0.8 & 0.8 & 0.9 & 0.8 & 0.9 & 0.9 \\
$ \gamma_e$ & 0.1 & 0.1 & 0.3 & 0.7 & 0.1 & 0.1 & 0.1 & 0.8 & 0.8 \\

\bottomrule 
\end{tabular}
}
\caption{Values for parameters $\gamma_r$ and $\gamma_e$ for CIFAR-10 and CIFAR-100.} 
\label{tab:parameters_cifar}
\end{table}

For CIFAR-10 and CIFAR-100 with IDN~\cite{xia2020part} we fixed $\gamma_r = 0.8$ and $\gamma_e = 0.1$ for all noise rates. We also use a WarmUp of 30 epochs (a WarmUp is a training using the whole training set to minimise the standard cross-entropy loss without applying any noisy-label learning technique), as suggested in~\cite{dividemix}. All other parameters were the same as described above for both CIFAR-10 and CIFAR-100. 
For the combined CIFAR-10 and CIFAR-100 experiment, we use $\gamma_r = 0.9$ and $\gamma_e = 0.1$ for all noise rates. All  other parameters are the same as described above for both CIFAR-10 and CIFAR-100.

For Webvision, we use InceptionResNet-V2 as backbone model, following~\citep{dividemix}.  We train the network with
SGD optimizer for 150 epochs with a momentum of 0.9 and a weight decay of 1e-4. The initial learning rate is 0.01, which gets  reduced by a factor of 10 after 50 and 100 epochs. The batchsize is fixed at 32. We used $\gamma_r=0.95$, as in ~\cite{ssr} and $\gamma_e=0.3$

 For ANIMAL-10N, we use VGG-19 as backbone model~\cite{koonce2021vgg}, which is trained with batch-normalization following~\cite{selfie}. We train the network with SGD optimizer for 150 epochs with a momentum of 0.9 and weight decay of 5e-4. The initial learning rate is 0.02 and reduced by a factor of 10 after 50 and 100 epochs. We used $\gamma_r=0.95$, as in~\cite{ssr}, and $\gamma_e=0.3$.

\section{Results}
\label{sec:results}

\subsection{Analysis of the proposed method}

\begin{figure}[!ht]
\centering
\includegraphics[ width=0.95\columnwidth]{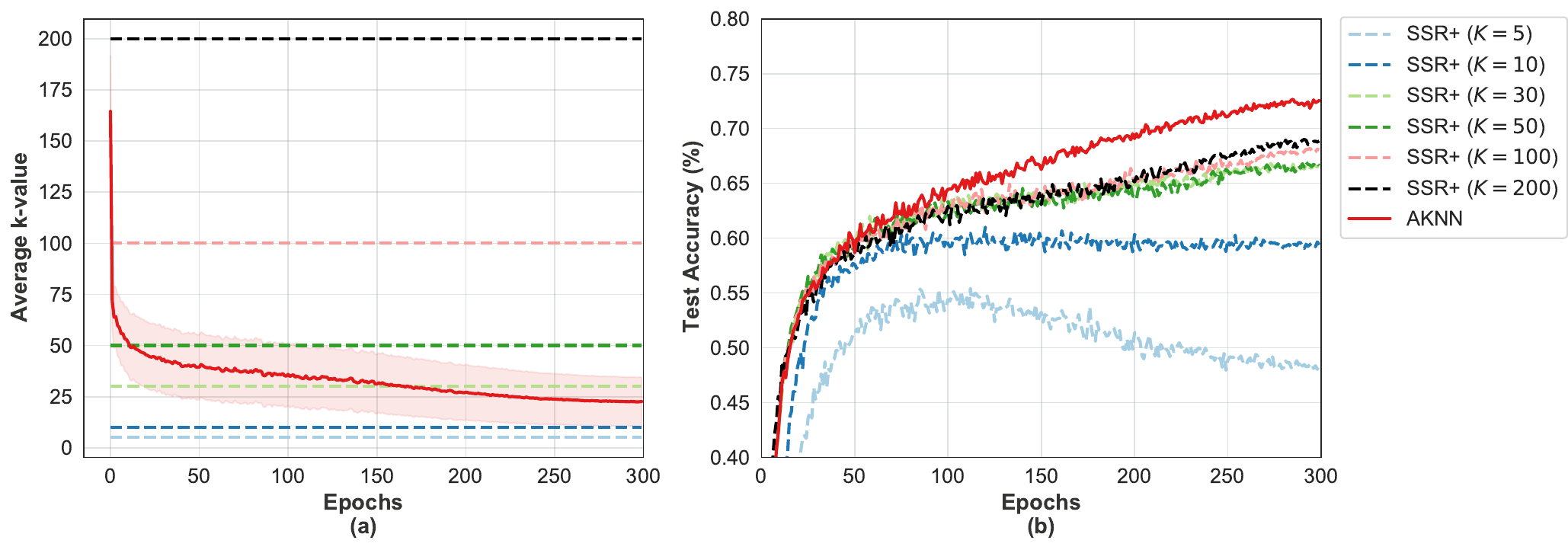}
\caption{Evaluation of SSR+~\cite{ssr} either using the original KNN (within SSR+ using $K \in \{5, 10, 30, 50, 100, 200\}$), or our proposed AKNN on CIFAR-100,  50\% symmetric noise: (a) average $K$-value through training epochs; (b) test accuracy results during  training.}
\label{fig:k_value}
\end{figure}

We first evaluate the effect of our proposed AKNN compared to the traditional KNN with the SSR+ framework~\cite{ssr} either using the original KNN or our proposed AKNN. 
For SSR+, while the baseline approach uses $K=200$ for all noise rates, in this study, we evaluate $K \in \{5, 10, 30, 50, 100, 200\}$ to show that our approach differs from just changing the neighbourhood size. 
For our AKNN, the $K$-value changes for each sample depending on the local feature density, and we compute the mean $K$-value of nearest neighbours during the training process. In the proposed AKNN method, we establish a minimum $K$-value based on $K_{LCS1}^{min}$ and $K_{LCS2}^{min}$, which are specifically defined for the $\mathcal{D}_{LCS}$ subset. For the $\mathcal{D}_{HCS}$ subset, we utilize the FINE approach within the ANNE framework. However, in the AKNN analysis, where the method is applied to the entire dataset for comparison with SSR+, we also use AKNN on $\mathcal{D}_{HCS}$, but with a reduced minimum $K$-value of 5, given that this subset consists of highly confident samples.
Fig.~\ref{fig:k_value}(a) shows that the average $K$-value in AKNN is initialised with a large value  that is subsequently reduced during training. At the beginning of the training, it is expected that samples have higher $K$-neighbours to decide if the sample is clean due to the initial uncertainty. As  training progresses and model training converges, AKNN adapts to search for a local neighborhood based on small values of $K$. Fig~\ref{fig:k_value}(b) shows that by simply substituting the original KNN strategy in SSR+ for the AKNN, the test accuracy  increases. 

\begin{figure}[!ht]
\centering
\includegraphics[ width=0.9\columnwidth]{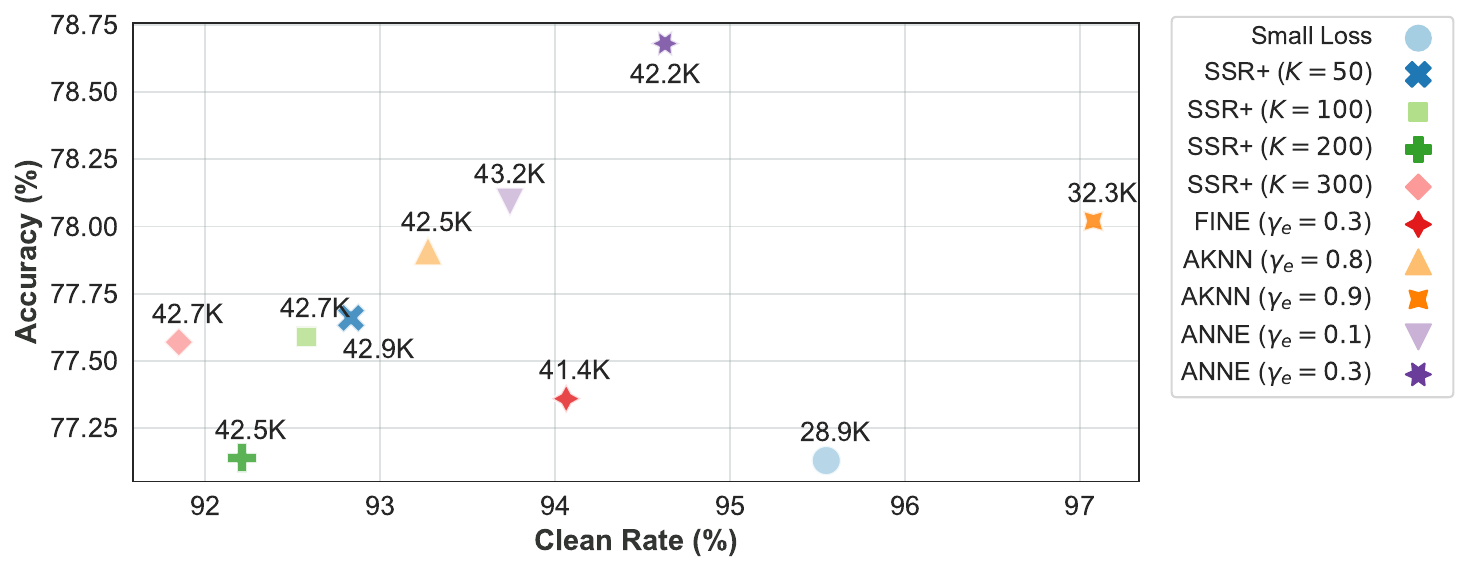}
\caption{Evaluation of sample selection size, clean label selection accuracy (clean rate) and accuracy on CIFAR-100, with 40\% instance dependent noisy. 
The graph compares the performance of ANNE (with parameter $\gamma_e = \{ 0.1,0.3 \}$) with SSR+~\cite{ssr} using the sample selection strategies of KNN (with parameter $K \in \{ 50,100,200,300 \}$), FINE (with parameter $\gamma_e = 0.3$), Small Loss, and AKNN (with parameter $\gamma_e = \{ 0.8,0.9 \}$).
The label on each marker refers to the size of the subset from the training set classified as clean label.
}
\label{fig:abl_ss}
\end{figure}

Next, we evaluate the sample selection size, clean label selection accuracy (clean rate) and accuracy of ANNE (with parameter $\gamma_e = \{ 0.1,0.3 \}$), compared with SSR+ with the sample selection strategies of KNN (with parameter $K \in \{ 50,100,200,300 \}$), FINE (with parameter $\gamma_e = 0.3$), Small Loss, and AKNN (with parameter $\gamma_e = \{ 0.8,0.9 \}$), all using the SSR+ framework~\cite{ssr}. The experiments are evaluated on CIFAR-100, 40\% instance-dependent noise. Fig.~\ref{fig:abl_ss} presents the results, indicating that methods with higher clean rates in the selected samples of the clean-label subset generally produce  a clean-label subset of  smaller  sizes. On the other hand, as the selection size increases, test accuracy also tends to increase. 
Hence, from the graph in Fig.~\ref{fig:abl_ss} it is clear that ANNE provides an advantageous tradeoff between clean rate and selection size, which positively correlates with higher accuracy.

\begin{figure}[!ht]
\centering
\includegraphics[ width=0.9\columnwidth]{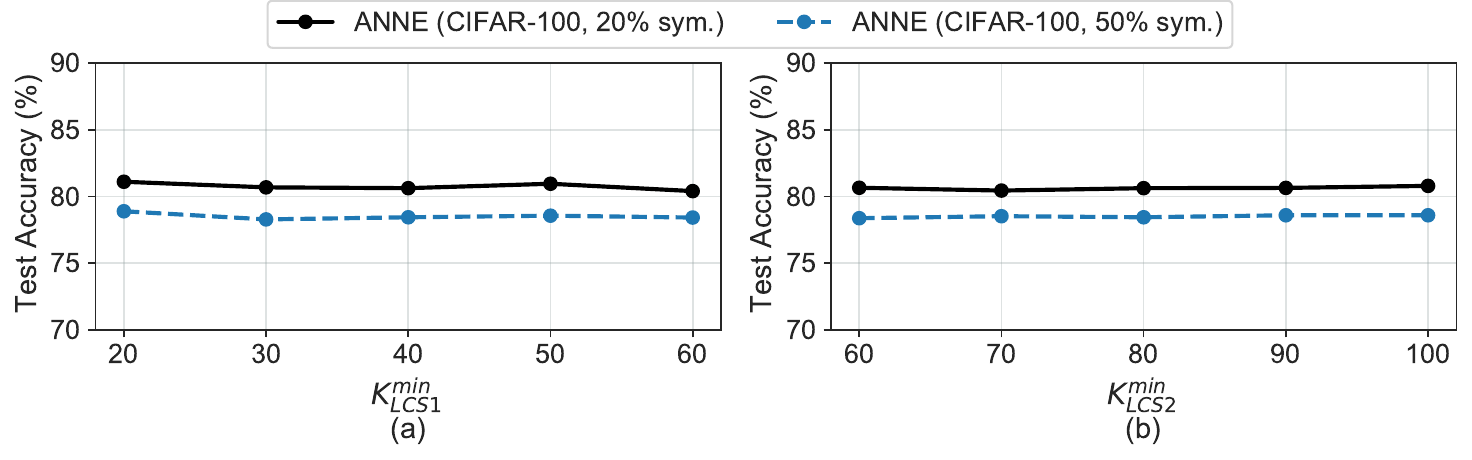}
\caption{Evaluation of robustness of our method with respect to $K_{LCS1}^{min}$ and $K_{LCS2}^{min}$ on CIFAR-100, with 20\% and 50\% symmetric noise. The graph compares the performance of ANNE with (a) $K_{LCS1}^{min}\in\{20, 30, 40, 50, 60\}$ and $K_{LCS2}^{min} = 80$, and (b) $K_{LCS1}^{min} = 40$ and $K_{LCS2}^{min} \in \{60, 70, 80, 90, 100\}$.
}
\label{fig:kmin}
\end{figure}

Next, we assess the robustness of the parameters $K_{LCS1}^{min}$ and $K_{LCS2}^{min}$. As described in Sec.~\ref{sec:AKNN}, we used fixed values of $K_{LCS1}^{min}=40$ and $K_{LCS2}^{min}=80$, for all the experiments. However, we show that different values of these parameters lead to similar results. Fig.~\ref{fig:kmin}~(a) presents the test accuracy of ANNE with  fixed $K_{LCS2}^{min} = 80$ and varying $K_{LCS1}^{min}=\{20, 30, 40, 50, 60\}$, whereas, Fig.~\ref{fig:kmin}~(b) shows  the test accuracy by fixing   $K_{LCS1}^{min} = 40$ and varying $K_{LCS2}^{min} = \{60, 70, 80, 90, 100\}$, for CIFAR-100, with 20\% and 50\% symmetric noise. The results demonstrate the robustness of ANNE with respect to these parameters across different noise levels.

\subsection{Comparison with State-of-the-Art}

We evaluate our model on CIFAR-10 and CIFAR-100 data sets using symmetric label noise ranging from 20\% to 90\% and 40\% asymmetric noise. We report the best test accuracy of all epochs, as commonly reported by methods in the literature, including the ones that we compare to in this section. Indeed, compared to SOTA methods, Table \ref{tab:results_cifar} shows  that our method obtains the most accurate classification results for almost all noise rates, obtaining large improvements on CIFAR-100.

\begin{table}[ht]
\centering
\footnotesize
\scalebox{0.88}{
\begin{tabular}{ll|cccc|c||cccl}
\toprule
\multicolumn{2}{c|}{\textbf{Data set}} & \multicolumn{5}{c||}{\textbf{CIFAR-10}} & \multicolumn{4}{c}{\textbf{CIFAR-100}}\\    
\cmidrule(l){1-2}
\cmidrule(l){3-7}
\cmidrule(l){8-11}
\multicolumn{2}{c|}{\textbf{Noise type}} & \multicolumn{4}{c}{\textbf{sym.}} & \multicolumn{1}{c||}{\textbf{asym.}} &  \multicolumn{4}{c}{\textbf{sym.}} \\
\midrule

Method & Sample Selection & 20\% & 50\% & 80\% & 90\% & 40\% & 20\% & 50\% & 80\% & 90\%  \\
\midrule
\multirow{1}{*}{Cross-Entropy \cite{dividemix}} & - & 86.8 & 79.4 & 62.9 & 42.7 & 85.0  & 62.0 & 46.7 & 19.9 & 10.1 \\
\multirow{1}{*}{Coteaching+ \cite{coteaching}} & Loss & 89.5 & 85.7 & 67.4& 47.9& - &65.6& 51.8 & 27.9 & 13.7\\
 
\multirow{1}{*}{ELR+~\cite{elr}}& - &95.8 & 94.8 & 93.3 & 78.7 & 93.0  & 77.6 & 73.6 & 60.8 & 33.4 \\
 
\multirow{1}{*}{DivideMix \cite{dividemix}}& Loss &96.1 & 94.6 & 93.2 & 76.0& 93.4  & 77.3 & 74.6 & 60.2 & 31.5 \\
\multirow{1}{*}{FINE~\cite{fine} }& Eigenvector &91.0 & 87.3 & 69.4 & - & 89.5  & 70.3 & 64.2 & 25.6 & -\\

\multirow{1}{*}{LongReMix \cite{longremix} }& Loss & 96.3 & 95.1 & 93.8 & 79.9 & 94.7  & 77.9 & 75.5 & 62.3 & 34.7\\
    
\multirow{1}{*}{Bootstrap~\cite{bootstrapping} }& Prediction &96.0 & 95.6 & 94.6 & 93.5 & 95.8  & 75.1 & 70.9 & 57.0 & 39.9\\

\multirow{1}{*}{SOP+~\cite{sop} }& - & 96.3 & 95.5 & 94.0 & - & 93.8  & 78.8 & 75.9 & 63.3 & -\\
\multirow{1}{*}{PES (semi)~\cite{pes} }& Prediction& 95.9 & 95.1 & 93.1 & - & -  & 77.4 & 74.3 & 61.6 & -\\
\multirow{1}{*}{C2D~\cite{c2d} }& Loss & 96.4 & 95.3 & 94.4 & 93.6 & 93.4  & 78.7 & 76.4 & 67.8 & 58.7\\
\multirow{1}{*}{DM-AugDesc
~\cite{nishi2021augmentation} }& Loss &96.3 & 95.6 & 93.7 & 35.3 & 94.4  & 79.6 & 76.6 & 61.8 & 17.3\\
\multirow{1}{*}{PropMix~\cite{propmix} }& Loss & 96.4 & 95.8 & 93.9 & 93.5 & 94.9  & 77.4 & 74.6 & 67.3 & 58.6\\
\multirow{1}{*}{MOIT~\cite{moit}} & KNN& 94.1 & 91.1 & 75.8 & 70.1 & 93.2  & 75.9 & 70.1 & 51.4 & 24.5\\

\multirow{1}{*}{ULC~\cite{ulc} }& Loss & 96.1 & 95.2 & 94.0 & 86.4 & 94.6  & 77.3 & 74.9 & 61.2 & 34.5\\

\multirow{1}{*}{SSR+~\cite{ssr} }& KNN & 96.7 &   96.1 & 95.6 & \textbf{95.2}  & 95.5 & 79.7 & 77.2 & 71.9 &\textbf{66.6} \\

\midrule

\multirow{1}{*}{ANNE (Ours) }& AKNN+Eigenv.& \textbf{96.9} & \textbf{96.2} & \textbf{95.8} & \textbf{95.3} & \textbf{95.7}  & \textbf{80.4} & \textbf{78.1} & \textbf{73.0} & 66.4\\
\bottomrule 
\end{tabular}
}
\caption{Test accuracy results (\%) for  all competing methods (discriminated by their noisy-label sample selection strategy) on CIFAR-10 and CIFAR-100 under symmetric (ranging from 20\% to 90\%) and asymmetric  noises.  Results of previous methods are the best test accuracies cited from their
original papers. Top methods are in bold.} 
\label{tab:results_cifar}
\end{table}

We also evaluate our approach on the CIFAR-10 and CIFAR-100 IDN benchmarks in Table~\ref{tab:results_idn}. Compared to SOTA approaches, we observe a large improvement of our approach on CIFAR-100 IDN. In fact, ANNE shows better results than SOTA in almost all cases on CIFAR-100 IDN. For CIFAR-10 IDN, results are quite similar to SOTA methods, which seem to have achieved a saturation point.

\begin{table}[ht]
\centering
\footnotesize
\scalebox{0.9}{
\begin{tabular}{ll|cccc||cccc}
\toprule
\multirow{2}{*}{Method} & \multirow{2}{*}{Sample Selection}& \multicolumn{4}{c||}{IDN - CIFAR10} & \multicolumn{4}{c}{IDN - CIFAR100}\\
\cmidrule(l){3-6}
\cmidrule(l){7-10}
 & & 0.2 & 0.3 & 0.4  & 0.5 & 0.2 & 0.3 & 0.4  & 0.5 \\
 \midrule
 CE~\cite{yao2021instance} & - & 75.81 & 69.15 & 62.45 & 
  39.42 & 30.42 & 24.15 &  15.23& 14.42 \\
 Mixup~\cite{mixup} & - & 73.17 & 70.02 & 61.56 & 
  48.95& 32.92 & 29.76 & 25.93 &  21.31 \\
 Forward~\cite{patrini2017making}& - & 74.64 & 69.75 & 
 60.21 & 46.27 & 36.38 & 33.17 & 26.75 & 19.27 \\
 T-Revision~\cite{xia2019anchor}& - & 76.15 & 70.36 & 
 64.09 &  49.02 & 37.24 & 36.54 & 27.23  & 22.54 \\
 
 Reweight~\cite{liu2015classification}& - & 76.23 & 
 70.12 & 62.58 & 45.46 &36.73 & 31.91 & 28.39 & 20.23 \\
 PTD-R-V~\cite{xia2020part}& - & 76.58 & 72.77 & 59.50 &
   56.32 & 65.33 & 64.56 & 59.73 &  56.80 \\
 Decoupling~\cite{decoupling}& Prediction & 78.71 & 75.17 & 61.73  & 
 50.43 & 36.53 & 30.93& 27.85  & 19.59 \\
 Co-teaching~\cite{coteaching}& Loss & 80.96 & 78.56 & 73.41 & 
  45.92 & 37.96 & 33.43 & 28.04 &  23.97 \\
 MentorNet~\cite{mentornet}& Loss &81.03 & 77.22 & 71.83 &
   47.89 & 38.91 & 34.23 & 31.89 & 24.15 \\
 CausalNL~\cite{yao2021instance}& Causal & 81.79 & 80.75 & 77.98 &
   78.63 & 41.47 & 40.98 & 34.02 &  32.13\\
 HOC~\cite{zhu2021clusterability}& KNN & 90.03 & - & 85.49 &
   - & 68.82 & -& 62.29 &  - \\
 CAL~\cite{zhu2021second}& Loss & 92.01 & - & 84.96 & 
  - & 69.11 & - & 63.17  & - \\
CC~\cite{centralityc}& Cosine. Sim. & 93.68 & - & 94.97 &  -  & 79.61 & - & 76.58 & - \\ 
 kMEIDTM~\cite{cheng2022instance}& Distillation & 92.26 & 90.73 & 85.94 &
   73.77 & 69.16 & 66.76 & 63.46 &  59.18 \\
 DivideMix~\cite{dividemix}& Loss & 94.80 & 94.60 & 94.53 & 
  93.04 & 77.07 & 76.33 & 70.80  & 58.61 \\
 InstanceGM~\cite{instancegm}& Loss & \textbf{96.68} & 96.52 & \textbf{96.36} &
   95.90 & 79.69 & \textbf{79.21} & 78.47 & 77.19 \\

 SSR+~\cite{ssr}& KNN& 96.46 & \textbf{96.61} & 96.14 & 
  96.18 & 78.66 & 78.66 & 77.30 &  76.03 \\
 \midrule
 ANNE (ours) & AKNN + Eigenv. & 96.43 & 96.32 & 96.20 & 
  \textbf{96.28} & \textbf{80.37} & 79.05 & \textbf{78.62}  & \textbf{77.35} \\
\bottomrule 
\end{tabular}
}
\caption{Test accuracy (\%) of different methods (discriminated by their noisy-label sample selection strategy) on CIFAR-10 and CIFAR-100  under various IDN noise rates. Results are extracted from \cite{instancegm}.
 Top methods are in bold.}  
\label{tab:results_idn}
\end{table}

We also evaluate our method for the combined open-set and symmetric closed-set noise on CIFAR-10 data set in Table~\ref{tab:res_combined}. Results show that our approach achieves competitive results when compared with SOTA methods. 

\begin{table}[!ht]
\footnotesize
\centering
\scalebox{1.0}{
\begin{tabular}{ll|c|c|c|c|c}
\toprule
\multirow{2}{*}{Method} & \multirow{2}{*}{Sample Selection} & Closed noise & 0.15 & 0 & 0.3 & 0\\
\cmidrule(l){3-7}
 &  & Open noise & 0.15 & 0.3 & 0.3 & 0.6 \\
 \midrule
 \multirow{2}{*}{ILON~\cite{ilon}} & KNN & Best & 87.4 & 90.4 & 80.5 & 83.4 \\
 & & Last & 80.0 & 87.4 & 55.2 & 78.0 \\
 \midrule
 \multirow{2}{*}{RoG~\cite{rog}} & Covariance & Best & 89.8 & 91.4 & 84.1 & 88.2 \\
  & Determinant &Last & 85.9 & 89.8 & 66.3 & 82.1 \\
  \midrule
\multirow{2}{*}{DivideMix~\cite{dividemix}} & \multirow{2}{*}{Loss} & Best & 91.5 & 89.3 & 91.8 & 89.0 \\
 & & Last & 90.9 & 88.7 & 91.5 & 88.7 \\
\midrule
\multirow{2}{*}{EDM~\cite{evidentialmix}}& Loss & Best & 94.5 & 92.9& 93.4 & 90.6 \\
 & & Last & 94.0 & 91.9 & 92.8 & 89.4 \\
\midrule
\multirow{2}{*}{SSR+~\cite{ssr}}&\multirow{2}{*}{KNN} & Best & 96.3 & 96.1 & \textbf{95.2}& \textbf{94.0} \\
& & Last & 96.2 & 96.0 & \textbf{95.2} & \textbf{93.9} \\
\midrule
\multirow{2}{*}{ANNE (ours)} & AKNN + &Best & \textbf{96.6} & \textbf{96.3} & \textbf{95.2} & 93.9 \\
 & Eigenvector & Last & \textbf{96.4}& \textbf{96.2} & 95.0 &93.7 \\
\bottomrule
\end{tabular}
}
\caption{Evaluation on CIFAR10 with combined closed (symmetric) and open-set noise. Results from baseline methods (discriminated by their noisy-label sample selection strategy) are as presented in \cite{ssr}. Top methods are in bold. }
\label{tab:res_combined}
\end{table}

The evaluation of ANNE on the large-scale real-world data set Webvision~\cite{webvision} is shown in Table~\ref{tab:res_WebVision}.
This table shows Top-1/-5 test accuracy on Webvision and ILSVRC12 test sets. 
Results show that ANNE is better than SOTA for all cases, which suggests that our approach is also effective in challenging large-scale data sets. 

\begin{table}[!ht]
\footnotesize
\centering
\scalebox{1.0}{
\begin{tabular}{ll|cc|cc}
\toprule
\multirow{2}{*}{Method}  & \multirow{2}{*}{Sample Selection}  &\multicolumn{2}{c|}{Webvision} & \multicolumn{2}{c}{ILSVRC12}  \\
   & & Top 1 & Top 5 & Top 1& Top 5  \\
\midrule
 D2L~\cite{d2l}            & - & 62.68 & 84.00  & 57.80 & 81.60  \\
 MentorNet~\cite{mentornet}  & Loss    & 63.00 & 81.40 & 57.80 & 79.92  \\
 Co-teaching~\cite{coteaching}& Loss    & 63.58 & 85.20 & 61.48& 84.70 \\
 Iterative-CV~\cite{chen2019understanding} & Prediction  & 65.24 & 85.34 & 61.60  & 84.98 \\
 DivideMix~\cite{dividemix}  & Loss    & 77.32 & 91.64 & 75.20& 90.84 \\
 ELR+~\cite{elr}   & -   & 77.78 & 91.68 &  70.29& 89.76 \\
 LongReMix~\cite{longremix}  & Loss   & 78.92  & 92.32 & - & -\\
 RRL~\cite{rrl} & KNN & 76.30& 91.50 & 73.30 & 91.20 \\
 CC~\cite{centralityc} & Cosine Similarity & 79.36 & 93.64 & 76.08  & \textbf{93.86} \\
 SSR+~\cite{ssr} & KNN & 80.92 & 92.80 & 75.76 & 91.76 \\
 NGC~\cite{ngc} & Prediction &  79.16 & 91.84 & 74.44 & 91.04 \\
 \midrule
 ANNE (ours) & KNN + Eigenvector & \textbf{81.96} & \textbf{93.96} & \textbf{76.80} & 92.68 \\
 \bottomrule 
 
\end{tabular}
}
\caption{Test Accuracy (\%) for  WebVision. Results from baseline methods (discriminated by their noisy-label sample selection strategy) are as presented in \cite{dividemix}. Top methods are in bold.}
\label{tab:res_WebVision}
\end{table}

Table~\ref{tab:res_animal} shows the results of our method on the real-world noisy-label ANIMAL-10N dataset. Results show that our method has competitive results compared with SOTA, but similarly to the CIFAR-10 IDN results in Table~\ref{tab:results_idn}, results on Animal-10N seem to have reached a saturation point.

\begin{table}[!ht]
\footnotesize
\centering
\scalebox{1.0}{
\begin{tabular}{ll|c}
\toprule
Method & Sample Selection & Test Accuracy (\%)\\
\hline
CE~\cite{zhang2021learning} & - &  79.4 \\
Nested-Dropout~\cite{chen2021boosting}& - & 81.3 \\
CE+Dropout~\cite{chen2021boosting}& - & 81.3 \\
SELFIE~\cite{selfie} & Label Consistency & 81.8 \\
PLC~\cite{zhang2021learning}& Prediction & 83.4 \\
Nested-CE~\cite{chen2021boosting}& Loss & 84.1 \\
InstanceGM~\cite{instancegm}& Loss & 84.6 \\
SSR+~\cite{ssr} & KNN & 88.5 \\
DivideMix~\cite{dividemix}& Loss & \textbf{88.8} \\
\midrule
ANNE (ours)& KNN + Eigenvector & 88.2 \\
\bottomrule
\end{tabular}
}
\caption{Test accuracy (\%) of different methods evaluated on
ANIMAL-10N. Results from baseline methods (discriminated by their noisy-label sample selection strategy) are as presented in \cite{ssr}. Top method is in bold.}
\label{tab:res_animal}
\end{table}

\subsection{Ablation Study}

We conduct an ablation study of ANNE on CIFAR-100 dataset under symmetric noise at various rates. Table~\ref{tab:abl_ours} presents a comparison of ANNE's performance with and without Adaptive KNN and FINE. We can observe that when ANNE is employed with FINE but without AKNN, it yields superior results at low noise rates. However, its performance diminishes at higher noise levels. On the other hand, using ANNE with AKNN but without FINE results in lower accuracy at low noise levels but demonstrates increased stability at higher noise rates. The combination of both strategies enables a more robust and consistent performance across both low and high noise rates.



\begin{table}[!ht]
\centering
\scalebox{0.8}{
\begin{tabular}{l|cccc}
\toprule
Ablation & 20\% & 50\% & 80\% & 90\%\\
\midrule
\textbf{ANNE} &  80.36 & 78.19 & 73.00 & 66.41  \\
ANNE w/ AKNN, w/o FINE & 79.23 & 77.33 & 72.66 & 64.37  \\
ANNE w/o AKNN, w/ FINE & 80.45 & 77.24 & 71.21 & 50.26  \\

\bottomrule
\end{tabular}
}
\caption{Ablation study of ANNE, showing the test accuracy (\%) of ANNE (original formulation), ANNE without FINE (i.e., with AKNN only), and ANNE without AKNN (i.e., with FINE only), on CIFAR-100, with symmetric noise at rates in $\{20\% , 50\% , 80\% , 90\% \}$.}
\label{tab:abl_ours}
\end{table}

We also study the use of FINE and AKNN on the different subsets 
$\mathcal{D}_{HCS},\mathcal{D}_{LCS}$
described in Sec.~\ref{sec:method}. Table~\ref{tab:abl_groups} shows these results on IDN CIFAR-100 dataset at noise rates 30\% and 50\%.
Notice in this table that the different setups exhibit similar results at low noise rates. However, as the noise rate increases, the combination setup using FINE in $\mathcal{D}_{HCS}$ and AKNN on $\mathcal{D}_{LCS}$ show the best results. 
We include an average ranking measure in the last column of Table~\ref{tab:abl_groups}, indicating that this setup ranks the highest.

\begin{table}[!ht]
\footnotesize
\centering
\scalebox{1.0}{
\begin{tabular}{cc|cc|c}
\toprule
\multicolumn{2}{c|}{Setup }& \multicolumn{2}{c|}{Acc} & \multirow{1}{*}{Mean Rank} \\
\cmidrule(l){1-4} 
 $\mathcal{D}_{HCS}$ & $\mathcal{D}_{LCS}$  & 30\% &  50\% &   \\
\midrule
FINE & FINE &  80.06 & 72.73 &  2.5 \\
FINE & AKNN &  79.75 &  77.35 &  \textbf{1.5}\\
AKNN & AKNN &  78.56 &  76.17 & 2.5\\
AKNN & FINE &  79.13 &  75.25 & 2.5\\ 
\bottomrule
\end{tabular}
}
\caption{Ablation study showing the test accuracy (\%) of ANNE, when it uses FINE and AKNN on different sample selection groups $\mathcal{D}_{HCS},\mathcal{D}_{LCS}$ on IDN CIFAR-100 dataset at rates $30\%$ and $50\%$. 
The last column shows the average ranking of each approach in these two problems (smaller is better).}
\label{tab:abl_groups}
\end{table}

\subsection{Training Time Comparison}

We compare the total training time for SSR+, FINE and ANNE. Since these models share the same backbone, their testing times are the same. We train all models on a machine equipped with a single GPU NVIDIA GeForce RTX 2080 Ti, with an Intel Core i7 processor. Table ~\ref{tab:time} presents the total training time on CIFAR-100 with symmetric noise rate of 50\%. The results indicate a small increase in training duration when using ANNE, with a 1.3 $\times$ longer training time compared to SSR+. This shows that ANNE not only provides improved accuracy results compared with SOTA approaches, but it also has a competitive training time with them.

\begin{table}[!ht]
\footnotesize
\centering
\scalebox{1.0}{
\begin{tabular}{cc}
\toprule
Method & Training Time (hours)\\
\midrule
SSR+ &  6.46 \\
FINE & 7.15\\
ANNE & 8.69\\ 
\bottomrule
\end{tabular}
}
\caption{Evaluation of the total training time (in hours) on CIFAR-100 with symmetric noise rate 50\% when training with ANNE, SSR+ and FINE.}
\label{tab:time}
\end{table}

\section{Conclusion}

We presented ANNE, a noisy label training algorithm with a novel hybrid adaptive strategy for selecting clean and noisy samples. Our approach shows that using different selection strategies for high and low-confidence samples can lead to more robust training, with a better balance between selection size and clean rate. Additionally, we introduced Adaptive KNN for noisy labels, which dynamically adapts to the learning stage and noise rate based on the local neighbourhood density. It is important to note that our method does not rely on pre-training, co-training or semi-supervised training, being competitive with SOTA. Our results on CIFAR-10/100, IDN benchmark, and WebVision and ANIMAL-10 outperform the SOTA methods in various datasets,  with improvements in high
noise problems. Regarding training time, results show that ANNE is competitive with SOTA.


This paper delves into a crucial aspect illustrated in Fig.~\ref{fig:abl_ss}: the intricate relationship among sample selection size, clean label selection accuracy, and model classification accuracy. Generally, a notable trend emerges, demonstrating a strong correlation between high clean label selection accuracy and elevated classification accuracy. However, in scenarios where clean label selection accuracy remains constant, a surge in sample selection size leads to a corresponding increase in classification accuracy. Consequently, the findings underscore that classification accuracy hinges on both a substantial sample selection size and a heightened clean label selection accuracy. This interesting discovery adds to the list of contributions of this paper.

The main limitation of the method lies in the specific hyper-parameters defined for each problem, as listed in Table~\ref{tab:parameters_cifar}.  In fact, the setting of problem-specific hyper-parameters is a common problem affecting all noisy-label learning SOTA methods~\cite{song2022learning}. We plan to develop an approach that automatically adapts the parameters for each problem. Another interesting problem uncovered by our paper is an issue present in virtually all noisy-label learning problems involving the CIFAR-10 dataset.  The issue is that recent SOTA approaches, including the proposed ANNE, are reaching between 95\% to 97\% classification accuracy, making the use of CIFAR-10 dataset questionable for the testing of future methods since these results indicate a saturation in classification accuracy. 

\section{Acknowledgments}

Supported by the Engineering and Physical Sciences Research Council (EPSRC) through grant EP/Y018036/1.

\bibliography{egbib}

\begin{thebibliography}{10}
\expandafter\ifx\csname url\endcsname\relax
  \def\url#1{\texttt{#1}}\fi
\expandafter\ifx\csname urlprefix\endcsname\relax\def\urlprefix{URL }\fi
\expandafter\ifx\csname href\endcsname\relax
  \def\href#1#2{#2} \def\path#1{#1}\fi

\bibitem{gad2018practical}
A.~F. Gad, Practical Computer Vision Applications Using Deep Learning with CNNs: With Detailed Examples in Python Using TensorFlow and Kivy, 1st Edition, Apress, Berkeley, 2018.

\bibitem{esteva2021deep}
A.~Esteva, K.~Chou, S.~Yeung, et~al., Deep learning-enabled medical computer vision, NPJ digital medicine 4~(1) (2021) 5.
\newblock \href {http://dx.doi.org/10.1038/s41746-020-00376-2} {\path{doi:10.1038/s41746-020-00376-2}}.

\bibitem{chai2021deep}
J.~Chai, H.~Zeng, A.~Li, E.~W.~T. Ngai, Deep learning in computer vision: A critical review of emerging techniques and application scenarios, Machine Learning with Applications 6 (2021) 100134.
\newblock \href {http://dx.doi.org/10.1016/j.mlwa.2021.100134} {\path{doi:10.1016/j.mlwa.2021.100134}}.

\bibitem{bansal2022systematic}
A.~Bansal, R.~Sharma, M.~Kathuria, A systematic review on data scarcity problem in deep learning: Solution and applications, ACM Computing Surveys 54 (2022) 1--29.
\newblock \href {http://dx.doi.org/10.1145/3502287} {\path{doi:10.1145/3502287}}.

\bibitem{liu2017online}
Y.~Liu, M.~Liu, An online learning approach to improving the quality of crowd-sourcing, IEEE/ACM Transactions on Networking 25 (2017) 2166--2179.
\newblock \href {http://dx.doi.org/10.1109/TNET.2017.2680245} {\path{doi:10.1109/TNET.2017.2680245}}.

\bibitem{frenay_classification_2014}
B.~Fr{\'e}nay, M.~Verleysen, Classification in the presence of label noise: A survey, IEEE Transactions on Neural Networks and Learning Systems 25 (2014) 845--869.
\newblock \href {http://dx.doi.org/10.1109/TNNLS.2013.2292894} {\path{doi:10.1109/TNNLS.2013.2292894}}.

\bibitem{zhang_understanding_2017}
C.~Zhang, S.~Bengio, M.~Hardt, B.~Recht, O.~Vinyals, Understanding deep learning (still) requires rethinking generalization, Communications of the ACM 64 (2021) 107--115.
\newblock \href {http://dx.doi.org/10.1145/3446776} {\path{doi:10.1145/3446776}}.

\bibitem{mentornet}
L.~Jiang, Z.~Zhou, T.~Leung, L.-J. Li, L.~Fei-Fei, {M}entor{N}et: Learning data-driven curriculum for very deep neural networks on corrupted labels, in: Proceedings of the 35th International Conference on Machine Learning, PMLR, 2018, pp. 2304--2313.

\bibitem{song2022learning}
H.~Song, M.~Kim, D.~Park, Y.~Shin, J.-G. Lee, Learning from noisy labels with deep neural networks: A survey, IEEE Transactions on Neural Networks and Learning Systems 34 (2023) 8135--8153.
\newblock \href {http://dx.doi.org/10.1109/TNNLS.2022.3152527} {\path{doi:10.1109/TNNLS.2022.3152527}}.

\bibitem{dividemix}
J.~Li, R.~Socher, S.~C. Hoi, Dividemix: Learning with noisy labels as semi-supervised learning, in: Proceedings of the 8th International Conference on Learning Representations, OpenReview.net, 2020.

\bibitem{selfie}
H.~Song, M.~Kim, J.-G. Lee, {SELFIE}: Refurbishing unclean samples for robust deep learning, in: Proceedings of the 36th International Conference on Machine Learning, PMLR, 2019.

\bibitem{zhang2018generalized}
Z.~Zhang, M.~R. Sabuncu, Generalized cross entropy loss for training deep neural networks with noisy labels, in: Proceedings of the 32nd International Conference on Neural Information Processing Systems, Curran Associates Inc., 2018.

\bibitem{ssr}
C.~Feng, G.~Tzimiropoulos, I.~Patras, {SSR}: An efficient and robust framework for learning with unknown label noise, in: Proceedings of the 33rd British Machine Vision Conference, {BMVA} Press, 2022.

\bibitem{longremix}
F.~R. Cordeiro, R.~Sachdeva, V.~Belagiannis, I.~Reid, G.~Carneiro, Longremix: Robust learning with high confidence samples in a noisy label environment, Pattern Recognition 133 (2023) 109013.
\newblock \href {http://dx.doi.org/10.1016/j.patcog.2022.109013} {\path{doi:10.1016/j.patcog.2022.109013}}.

\bibitem{wei2020combating}
H.~Wei, L.~Feng, X.~Chen, B.~An, Combating noisy labels by agreement: A joint training method with co-regularization, in: Proceedings of the IEEE/CVF Conference on Computer Vision and Pattern Recognition, IEEE, 2020.

\bibitem{coteaching}
B.~Han, Q.~Yao, X.~Yu, G.~Niu, M.~Xu, W.~Hu, I.~Tsang, M.~Sugiyama, Co-teaching: Robust training of deep neural networks with extremely noisy labels, in: Advances in Neural Information Processing Systems, Curran Associates, Inc., 2018.

\bibitem{fine}
T.~Kim, J.~Ko, S.~Cho, J.~Choi, S.-Y. Yun, Fine samples for learning with noisy labels, in: Proceedings of the 35th International Conference on Neural Information Processing Systems, Curran Associates Inc., 2021.

\bibitem{larose2014k}
D.~T. Larose, C.~D. Larose, K-nearest neighbor algorithm, in: Discovering Knowledge in Data: An Introduction to Data Mining, Wiley, 2014.

\bibitem{cifar}
A.~Krizhevsky, G.~Hinton, Learning multiple layers of features from tiny images, Technical Report TR-2009, University of Toronto (2009).

\bibitem{shen2019learning}
Y.~Shen, S.~Sanghavi, Learning with bad training data via iterative trimmed loss minimization, in: Proceedings of the 36th International Conference on Machine Learning, PMLR, 2019.

\bibitem{kaiser2022blind}
T.~Kaiser, L.~Ehmann, C.~Reinders, B.~Rosenhahn, \href{https://arxiv.org/abs/2211.11355}{Blind knowledge distillation for robust image classification} (2022).
\newblock \href {http://arxiv.org/abs/2211.11355} {\path{arXiv:2211.11355}}.
\newline\urlprefix\url{https://arxiv.org/abs/2211.11355}

\bibitem{propmix}
F.~R. Cordeiro, V.~Belagiannis, I.~Reid, G.~Carneiro, Propmix: Hard sample filtering and proportional mixup for learning with noisy labels, in: Proceedings of the 32nd British Machine Vision Conference, {BMVA} Press, 2021.

\bibitem{o2unet}
J.~Huang, L.~Qu, R.~Jia, B.~Zhao, O2u-net: A simple noisy label detection approach for deep neural networks, in: Proceedings of the 2019 IEEE/CVF International Conference on Computer Vision, 2019.

\bibitem{wu2020topological}
P.~Wu, S.~Zheng, M.~Goswami, D.~Metaxas, C.~Chen, A topological filter for learning with label noise, in: Proceedings of the 34th International Conference on Neural Information Processing Systems, Curran Associates Inc., 2020.

\bibitem{wu2021ngc}
Z.-F. Wu, T.~Wei, J.~Jiang, C.~Mao, M.~Tang, Y.-F. Li, Ngc: A unified framework for learning with open-world noisy data, in: Proceedings of the IEEE/CVF International Conference on Computer Vision, IEEE, 2021.

\bibitem{centralityc}
G.~Zhao, G.~Li, Y.~Qin, F.~Liu, Y.~Yu, Centrality and consistency: Two-stage clean samples identification for learning with instance-dependent noisy labels, in: Proceeedings of the European Conference on Computer Vision, Springer, 2022.

\bibitem{kim2019nlnl}
Y.~Kim, J.~Yim, J.~Yun, J.~Kim, Nlnl: Negative learning for noisy labels, in: Proceedings of the IEEE International Conference on Computer Vision, IEEE, 2019.

\bibitem{patrini2017making}
G.~Patrini, A.~Rozza, A.~Krishna~Menon, R.~Nock, L.~Qu, Making deep neural networks robust to label noise: A loss correction approach, in: Proceedings of the IEEE/CVF Conference on Computer Vision and Pattern Recognition, IEEE, 2017.

\bibitem{rog}
K.~Lee, S.~Yun, K.~Lee, H.~Lee, B.~Li, J.~Shin, Robust inference via generative classifiers for handling noisy labels, in: International Conference on Machine Learning, PMLR, 2019.

\bibitem{otsu}
N.~Otsu, A threshold selection method from gray-level histograms, IEEE Transactions on Systems, Man, and Cybernetics 9 (1979) 62--66.
\newblock \href {http://dx.doi.org/10.1109/TSMC.1979.4310076} {\path{doi:10.1109/TSMC.1979.4310076}}.

\bibitem{mixup}
H.~Zhang, M.~Cisse, Y.~N. Dauphin, D.~Lopez-Paz, Mixup: Beyond empirical risk minimization, in: Proceedings of the 6th International Conference on Learning Representations, OpenReview.net, 2018.

\bibitem{webvision}
W.~Li, L.~Wang, W.~Li, E.~Agustsson, L.~Van~Gool, Webvision database: Visual learning and understanding from web data, in: Proceedings of the IEEE/CVF Conference on Computer Vision and Pattern Recognition, 2017.

\bibitem{xia2020part}
X.~Xia, T.~Liu, B.~Han, N.~Wang, M.~Gong, H.~Liu, G.~Niu, D.~Tao, M.~Sugiyama, Part-dependent label noise: Towards instance-dependent label noise, in: Advances in Neural Information Processing Systems, Curran Associates, Inc., 2020.

\bibitem{evidentialmix}
R.~Sachdeva, F.~R. Cordeiro, V.~Belagiannis, I.~Reid, G.~Carneiro, Evidentialmix: Learning with combined open-set and closed-set noisy labels, in: Proceedings of the IEEE/CVF Winter Conference on Applications of Computer Vision, 2021.

\bibitem{deng2009imagenet}
J.~Deng, W.~Dong, R.~Socher, L.-J. Li, K.~Li, L.~Fei-Fei, Imagenet: A large-scale hierarchical image database, in: Proceedings of the IEEE/CVF Conference on Computer Vision and Pattern Recognition, 2009.

\bibitem{chen2019understanding}
P.~Chen, B.~B. Liao, G.~Chen, S.~Zhang, Understanding and utilizing deep neural networks trained with noisy labels, in: Proceedings of the 36th International Conference on Machine Learning, PMLR, 2019.

\bibitem{he2016identity}
K.~He, X.~Zhang, S.~Ren, J.~Sun, Identity mappings in deep residual networks, in: European Conference on Computer Vision, Springer, 2016.

\bibitem{koonce2021vgg}
B.~Koonce, Convolutional Neural Networks with Swift for TensorFlow: Image Recognition and Dataset Categorization, 1st Edition, Apress, 2021.

\bibitem{elr}
S.~Liu, J.~Niles-Weed, N.~Razavian, C.~Fernandez-Granda, Early-learning regularization prevents memorization of noisy labels, in: Advances in Neural Information Processing Systems, Curran Associates, Inc., 2020.

\bibitem{bootstrapping}
B.~Smart, G.~Carneiro, Bootstrapping the relationship between images and their clean and noisy labels, in: Proceedings of the IEEE/CVF Winter Conference on Applications of Computer Vision, IEEE, 2023.

\bibitem{sop}
S.~Liu, Z.~Zhu, Q.~Qu, C.~You, Robust training under label noise by over-parameterization, in: International Conference on Machine Learning, PMLR, 2022.

\bibitem{pes}
Y.~Bai, E.~Yang, B.~Han, Y.~Yang, J.~Li, Y.~Mao, G.~Niu, T.~Liu, Understanding and improving early stopping for learning with noisy labels, in: Proceedings of the 35th International Conference on Neural Information Processing Systems, Curran Associates Inc., 2021.

\bibitem{c2d}
E.~Zheltonozhskii, C.~Baskin, A.~Mendelson, A.~M. Bronstein, O.~Litany, Contrast to divide: Self-supervised pre-training for learning with noisy labels, in: Proceedings of the IEEE/CVF Winter Conference on Applications of Computer Vision, IEEE, 2022.

\bibitem{nishi2021augmentation}
K.~Nishi, Y.~Ding, A.~Rich, T.~Hollerer, Augmentation strategies for learning with noisy labels, in: Proceedings of the IEEE/CVF Conference on Computer Vision and Pattern Recognition, IEEE, 2021.

\bibitem{moit}
D.~Ortego, E.~Arazo, P.~Albert, N.~E. O'Connor, K.~McGuinness, Multi-objective interpolation training for robustness to label noise, in: Proceedings of the IEEE/CVF Conference on Computer Vision and Pattern Recognition, IEEE, 2021.

\bibitem{ulc}
Y.~Huang, B.~Bai, S.~Zhao, K.~Bai, F.~Wang, Uncertainty-aware learning against label noise on imbalanced datasets, in: Proceedings of the AAAI Conference on Artificial Intelligence, AAAI Press, 2022.

\bibitem{yao2021instance}
Y.~Yao, T.~Liu, M.~Gong, B.~Han, G.~Niu, K.~Zhang, Instance-dependent label-noise learning under structural causal models, in: Proceedings of the 35th International Conference on Neural Information Processing Systems, Curran Associates Inc., 2021.

\bibitem{xia2019anchor}
X.~Xia, T.~Liu, N.~Wang, B.~Han, C.~Gong, G.~Niu, M.~Sugiyama, Are anchor points really indispensable in label-noise learning?, in: Advances in Neural Information Processing Systems, Curran Associates, Inc., 2019.

\bibitem{liu2015classification}
T.~Liu, D.~Tao, Classification with noisy labels by importance reweighting, IEEE Transactions on Pattern Analysis and Machine Intelligence 38 (2016) 447--461.
\newblock \href {http://dx.doi.org/10.1109/TPAMI.2015.2456899} {\path{doi:10.1109/TPAMI.2015.2456899}}.

\bibitem{decoupling}
E.~Malach, S.~Shalev-Shwartz, Decoupling "when to update" from "how to update", in: Advances in Neural Information Processing Systems, Curran Associates, Inc., 2017.

\bibitem{zhu2021clusterability}
Z.~Zhu, Y.~Song, Y.~Liu, Clusterability as an alternative to anchor points when learning with noisy labels, in: Proceedings of the 38th International Conference on Machine Learning, PMLR, 2021.

\bibitem{zhu2021second}
Z.~Zhu, T.~Liu, Y.~Liu, A second-order approach to learning with instance-dependent label noise, in: Proceedings of the IEEE/CVF Conference on Computer Vision and Pattern Recognition, IEEE, 2021.

\bibitem{cheng2022instance}
D.~Cheng, T.~Liu, Y.~Ning, N.~Wang, B.~Han, G.~Niu, X.~Gao, M.~Sugiyama, Instance-dependent label-noise learning with manifold-regularized transition matrix estimation, in: Proceedings of the IEEE/CVF Conference on Computer Vision and Pattern Recognition, IEEE, 2022.

\bibitem{instancegm}
A.~Garg, C.~Nguyen, R.~Felix, T.-T. Do, G.~Carneiro, Instance-dependent noisy label learning via graphical modelling, in: Proceedings of the IEEE/CVF Winter Conference on Applications of Computer Vision, IEEE, 2023.

\bibitem{ilon}
Y.~Wang, W.~Liu, X.~Ma, J.~Bailey, H.~Zha, L.~Song, S.-T. Xia, Iterative learning with open-set noisy labels, in: Proceedings of the IEEE/CVF Conference on Computer Vision and Pattern Recognition, IEEE, 2018.

\bibitem{d2l}
X.~Ma, Y.~Wang, M.~E. Houle, S.~Zhou, S.~Erfani, S.~Xia, S.~Wijewickrema, J.~Bailey, Dimensionality-driven learning with noisy labels, in: Proceedings of the 35th International Conference on Machine Learning, PMLR, 2018.

\bibitem{rrl}
J.~Li, C.~Xiong, S.~C. Hoi, Learning from noisy data with robust representation learning, in: Proceedings of the IEEE/CVF International Conference on Computer Vision, IEEE, 2021.

\bibitem{ngc}
Z.-F. Wu, T.~Wei, J.~Jiang, C.~Mao, M.~Tang, Y.-F. Li, Ngc: A unified framework for learning with open-world noisy data, in: Proceedings of the IEEE/CVF International Conference on Computer Vision, IEEE, 2021.

\bibitem{zhang2021learning}
Y.~Zhang, S.~Zheng, P.~Wu, M.~Goswami, C.~Chen, Learning with feature-dependent label noise: {A} progressive approach, in: Proceedings of the 9th International Conference on Learning Representations, ICLR, 2021.

\bibitem{chen2021boosting}
Y.~Chen, X.~Shen, S.~Hu, J.~Suykens, Boosting co-teaching with compression regularization for label noise, in: Conference on Computer Vision and Pattern Recognition Workshops, IEEE, 2021.

\end{thebibliography}

\newpage

\begin{minipage}{0.3\textwidth}
\begin{tikzpicture}[remember picture, overlay]
\node[text width=\linewidth]{
\begin{figure}[H]\centering
    \vspace{-2cm}
    \includegraphics[height=18cm, width=0.7\linewidth, keepaspectratio]{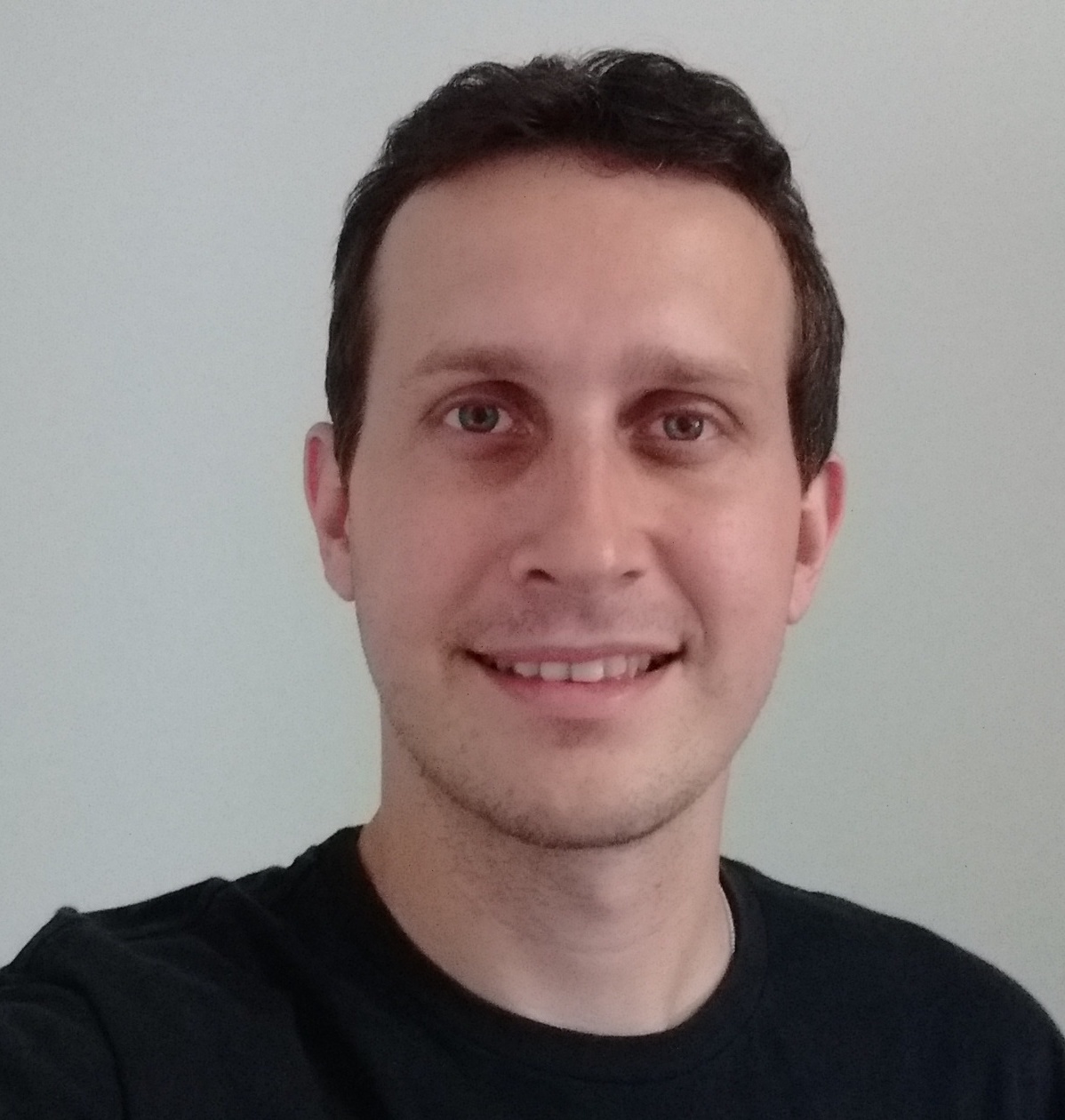}\\
    \vspace{1cm}
    
    \vspace{1cm}
    \includegraphics[height=18cm, width=0.7\linewidth, keepaspectratio]{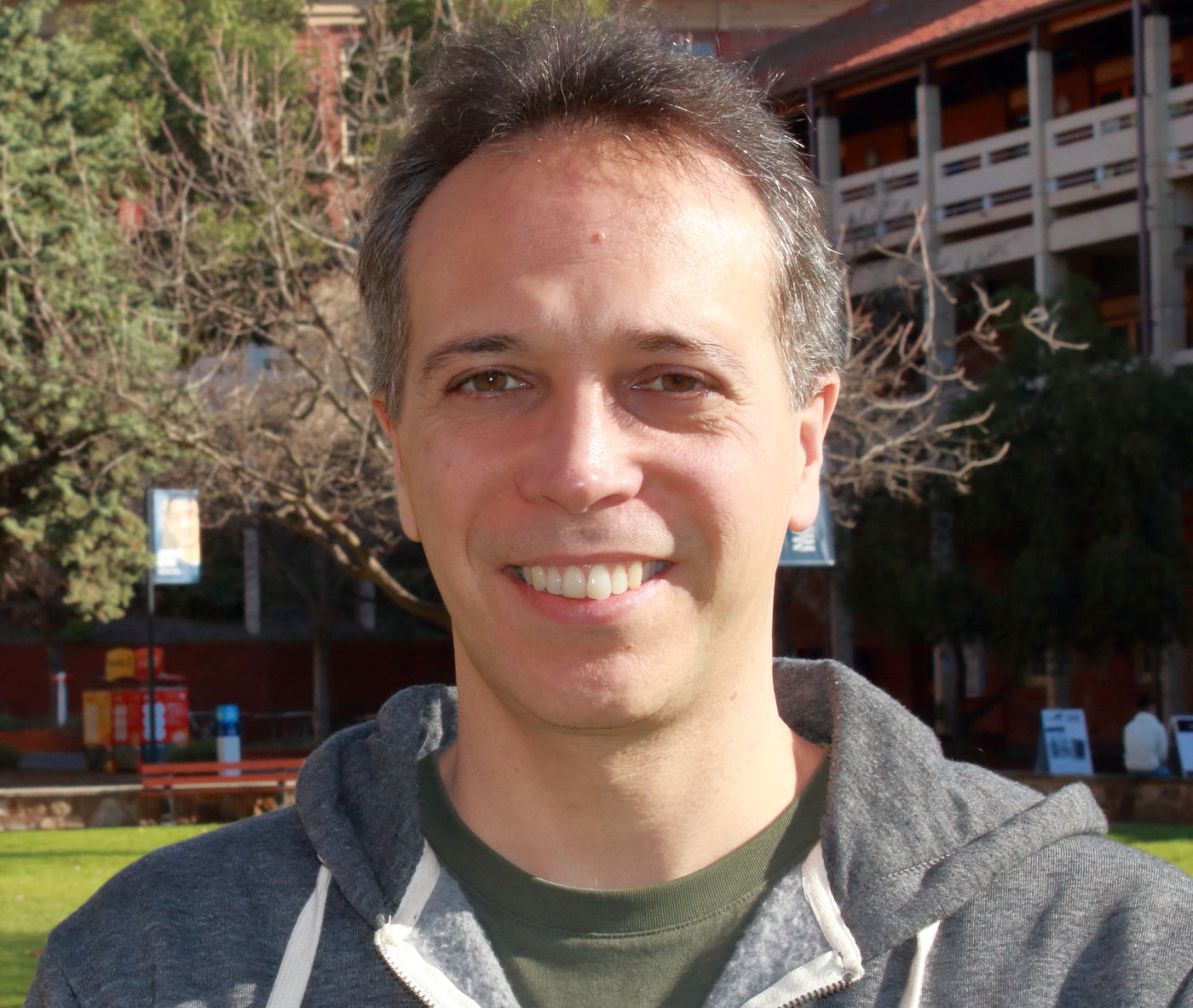}
\end{figure}
};
\end{tikzpicture}
\end{minipage}%
\hfill
\begin{minipage}{0.60\textwidth}\fontsize{8pt}{12pt}\selectfont
\begin{tabular}{p{\textwidth}}
\thispagestyle{empty}
Filipe R. Cordeiro is a professor of the Department of Computing at Universidade Federal Rural de Pernambuco (UFRPE). 
In 2015, he received his PhD in computer science from the Federal University of Pernambuco (UFPE). 
Filipe's main contributions are in the area of computer vision, medical image analysis, and machine learning.

\\[1\baselineskip]
Gustavo Carneiro is a professor in the Centre for Vision, Speech and Signal Processing at the University of Surrey and Lead of the People-centred Mammogram Analysis project (PecMan). His main research interests are in computer vision, medical image analysis and machine learning.

\end{tabular}
\end{minipage}%

\end{document}